\documentclass[11pt]{article} 
\usepackage{amsfonts, amssymb, amsmath}
\usepackage{algorithm, algorithmic, amsthm}
\usepackage{graphicx}
\usepackage{color}
\usepackage{hyperref}
\usepackage[top=2.5cm, bottom=3cm, left=3cm, right=3cm]{geometry}

\usepackage{subfigure}

\usepackage{times}
\usepackage{hyperref}
\usepackage{url}
\usepackage{graphicx, wrapfig}
\usepackage{amsfonts, amssymb, amsmath}
\usepackage{algorithm, algorithmic, amsthm}
\usepackage{subfigure}
\usepackage{multirow}
\usepackage{soul}
\usepackage{enumitem}
\usepackage{color}
\usepackage{booktabs}
\usepackage{upgreek}
\usepackage[hang]{footmisc}

\newcommand{\HH}{\ensuremath{\mathbf{H}}}

\newcommand{\M}{\ensuremath{\mathbf{M}}}

\newcommand{\h}{\ensuremath{\mathbf{h}}}

\newcommand{\rr}{\ensuremath{\mathbf{r}}}
\newcommand{\sss}{\ensuremath{\mathbf{s}}}  

\newcommand{\uu}{\ensuremath{\mathbf{u}}}

\newcommand{\x}{\ensuremath{\mathbf{x}}}

\newcommand{\calD}{\ensuremath{\mathcal{D}}}

\newcommand{\calT}{\ensuremath{\mathcal{T}}}

{%
\begin{list}{#1}{
\vspace{-\topsep}
\vspace{-\partopsep}
\setlength{\itemindent}{0cm}
\setlength{\rightmargin}{0cm}
\setlength{\listparindent}{0cm}
\settowidth{\labelwidth}{#1}
\setlength{\leftmargin}{\labelwidth}
\addtolength{\leftmargin}{\labelsep}
\setlength{\itemsep}{0cm}
}%
}%
{%
\end{list}
\vspace{-\topsep}
\vspace{-\partopsep}
}

{\begin{enumerate}%
}%
{\end{enumerate}}

\theoremstyle{plain}

\newtheorem*{lemma*}{Lemma}

\newtheorem*{prop*}{Proposition}

\theoremstyle{definition}

\newtheorem*{defn*}{Definition}

\newtheorem*{exmp*}{Example}

\newtheorem*{conj*}{Conjecture}

\theoremstyle{remark}

\newtheorem*{rmk*}{Remark}

\title{Neural Generative Question Answering}
\author{\sf Jun Yin$^1$\thanks{The work is done when the first author worked as intern at Noah's Ark Lab, Huawei Technologies.} \ \;
Xin Jiang$^2$\; Zhengdong Lu$^2$\; \\
\sf Lifeng Shang$^2$\; Hang Li$^2$ \; Xiaoming Li$^{1}$\\
$^1$School of Electronic Engineering and Computer Science, Peking University\\ {\tt \{jun.yin, lxm\}@pku.edu.cn}\\
$^2$Noah's Ark Lab, Huawei Technologies\\ {\tt \{Jiang.Xin, Lu.Zhengdong, Shang.Lifeng, HangLi.HL\}@huawei.com}}
\date{}

\begin{document}

\maketitle

\begin{abstract}
	This paper presents an end-to-end neural network model, named Neural Generative Question Answering (\textsc{genQA}), that can generate answers to \textit{simple factoid questions}, based on the facts in a knowledge-base. More specifically, the model is built on the encoder-decoder framework for sequence-to-sequence learning, while equipped with the ability to enquire the knowledge-base, and is trained on a corpus of question-answer pairs, with their associated triples in the knowledge-base. Empirical study shows the proposed model can effectively deal with the variations of questions and answers, and generate right and natural answers by referring to the facts in the knowledge-base. The experiment on question answering demonstrates that the proposed model can outperform an embedding-based QA model as well as a neural dialogue model trained on the same data.
\end{abstract}

\section{Introduction}\label{sec:introduction}
Question answering (QA) can be viewed as a special case of single-turn dialogue: QA aims at providing correct answers to the questions in natural language, while dialogue emphasizes on generating relevant and fluent responses to the messages also in natural language~\cite{shang15neural,vinyals2015neural}. Recent progress in deep learning has raised the possibility of realizing generation-based QA in a purely neutralized way. That is, the answer is generated by a neural network (e.g., recurrent neural network, or RNN) based on the question, which is able to handle the flexibility and diversity of language. More importantly, the model is trained in an end-to-end fashion, and thus there is no need in building the system using linguistic knowledge, e.g., creating a semantic parser.

There is however one serious limitation of this generation-based approach to QA. It is practically impossible to store all the knowledge in a neural network to achieve a desired precision and coverage in real world QA. This is a fundamental difficulty, rooting deeply in the way in which knowledge is acquired, represented and stored. The neural network, and more generally the fully distributed way of representation, is good at representing smooth and shared patterns, i.e., modeling the flexibility and diversity of language, but improper for representing discrete and isolated concepts, i.e., depicting the lexicon of language.

On the other hand, the recent success of memory-based neural network models has greatly extended the ways of storing and accessing text information, in both short-term memory (e.g., in~\cite{bahdanau2014neural}) and long-term memory (e.g., in~\cite{memorynet}). It is hence a natural choice to connect a neural model for QA with a neural model of knowledge-base on an external memory, which is also related to the traditional approach of template-based QA from knowledge-base.

In this paper, we report our exploration in this direction, with a proposed model called \emph{Neural Generative Question Answering} (\textsc{genQA}). The model can generate answers to \textit{simple factoid questions} by accessing a knowledge-base. More specifically, the model is built on the encoder-decoder framework for sequence-to-sequence learning, while equipped with the ability to enquire a knowledge-base. Its specifically designed decoder, controlled by another neural network, can switch between generating a common word (e.g., \texttt{is}) and outputting a term (e.g., ``\texttt{John Malkovich}" ) retrieved from knowledge-base with a certain probability. The model is trained on a dataset composed of real world question-answer pairs associated with triples in the knowledge-base, in which all components of the model are jointly tuned. Empirical study shows the proposed model can effectively capture the variation of language and generate right and natural answers to the questions by referring to the facts in the knowledge-base. The experiment on question answering demonstrates that the proposed model can outperform an embedding-based QA model as well as a neural dialogue model trained on the same data.

\section{Task Description}\label{sec:learning_task}

\begin{table*}
	\centering
	\small
	\caption{Examples of training instances for generative QA. The KB-words in the training instances are underlined in the examples. }\label{tab:examples}
	\begin{tabular}{p{9cm} p{6cm}}
		\hline
		Question \& Answer & Triple (\textit{subject}, \textit{predicate}, \textit{object})\\
		\hline
		Q: \textit{How tall is Yao Ming?} \newline A: \textit{He is \underline{2.29m} and is visible from space.} & (\texttt{Yao Ming, height, 2.29m}) \\
		\hline
		Q: \textit{Which country was Beethoven from?} \newline A: \textit{He was born in what is now \underline{Germany}.} & (\texttt{Ludwig van Beethoven, place of birth, Germany})\\
		\hline
		Q: \textit{Which club does Messi play for?} \newline A: \textit{Lionel Messi currently plays for \underline{FC Barcelona} in the Spanish Primera Liga.} &(\texttt{Lionel Messi, team, FC Barcelon})\\
		\hline
	\end{tabular}
\end{table*}

\subsection{The learning task}
We formalize generative question answering as a supervised learning task or more specifically a sequence-to-sequence learning task. A generative QA system takes a sequence of words as input question and generates another sequence of words as output answer. In order to provide right answers, the system is connected with a knowledge-base (KB), which contains facts. During the process of answering, the system queries the KB, retrieves a set of candidate facts and generates a correct answer to the question using the right fact. The generated answer may contain two types of ``words": one is common words for composing the answer (referred to as common word) and the other is specialized words in the KB denoting the answer (referred to as KB-word).

To learn a model for the task, we assume that each training instance consists of a question-answer pair with the KB-word specified in the answer. In this paper, we only consider the case of \textit{simple factoid question}, which means each question-answer pair is associated with a single fact (i.e., one triple) of the KB. Without loss of generality, we focus on forward relation QA, where the question is on \textit{subject} and \textit{predicate} of the triple and the answer is from \textit{object}. Tables~\ref{tab:examples} shows some examples of the training instances.

\subsection{Data}
To facilitate research on the task of generative QA, we create a new dataset by collecting data from the web. We first build a knowledge-base by mining from three Chinese encyclopedia web sites\footnote{Baidu Baike, Baike.com, Douban.com}. Specifically we extract entities and associated triples (\textit{subject}, \textit{predicate}, \textit{object}) from the structured parts (e.g. HTML tables) of the web pages at the web sites. Then the extracted data is normalized and aggregated to form a knowledge-base. In this paper we sometimes refer to an item of a triple as a constituent of knowledge-base. Second, we collect question-answer pairs by extracting from two Chinese community QA sites\footnote{Baidu Zhidao, Sogou Wenwen}. Table~\ref{tab:data_stat} shows the statistics of the knowledge-base and QA-pairs.

\begin{table}
	\centering
	\caption{Statistics of the QA data and the knowledge-base.}\label{tab:data_stat}
	\begin{tabular}{c | c  c }
		\hline
		Community QA & \multicolumn{2}{c}{Knowledge-base} \\
		\#QA pairs & \#entities & \#triples \\
		\hline
		235,171,463 & 8,935,028 & 11,020,656  \\
		\hline
	\end{tabular}
\end{table}

We automatically and heuristically construct training and test data for generative QA by ``grounding'' the QA pairs with the triples in the knowledge-base. Specifically, for each QA pair, a list of candidate triples with the \textit{subject} fields appearing in the question, is retrieved by using the Aho-Corasick string searching algorithm. The triples in the candidate list are then judged by a series of rules for relevance to the QA pair. The basic requirement for relevance is that the answer contains the \textit{object} of the triple, which specifies the KB-word in the answer. Besides, we use additional scoring and filtering rules, attempting to find out the triple that truly matches the QA pair, if there is any. As the result of processing, 720K instances (tuples of question, answer, triple) are finally obtained with an estimated accuracy of 80\%, i.e., 80\% of instances have truly correct grounding. The data is publicly available online\footnote{\url{https://github.com/jxfeb/Generative_QA}}.

The data is further randomly partitioned into training dataset and test dataset by using triple as the partition key.  In this way, all the questions in the test data are regarding to the unseen facts (triples) in the training data. Table~\ref{tab:training_data_stat} shows some statistics of the datasets. By comparing the numbers of triples in Table~\ref{tab:data_stat} and Table~\ref{tab:training_data_stat}, we can see that a large portion of facts in the knowledge-base are not present in the training and test data, which demonstrates the necessity for the system to generalize to unseen facts.

\subsection{Challenges}
The key challenge in learning of generative QA is to find a way to jointly train the neural network model in order to conduct understanding of question, generation of answer, and retrieval of relevant facts in KB, in a single and unified framework. To make things even harder, the data for training is noisy and informal, with typos, nonstandard expressions, and a wide range of language variations, which can block the system to acquire the right question-answer patterns.

\begin{table}
	\centering
	\caption{Statistics of the training and test dataset for \textsc{genQA}}\label{tab:training_data_stat}
	\begin{tabular}{ c c | c c}
		\hline
		\multicolumn{2}{c|}{Training Data} & \multicolumn{2}{c}{Test Data} \\
		\#QA pairs & \#triples & \#QA pairs & \#triples \\
		\hline
		696,306 & 58,019 & 23,364 & 1,974 \\
		\hline
	\end{tabular}
\end{table}

\section{The \textsc{genQA} Model} \label{sec:model}

\begin{figure}[htc]
	\begin{center}
		\begin{tabular}[c]{cc}
			\includegraphics[width=0.55\textwidth]{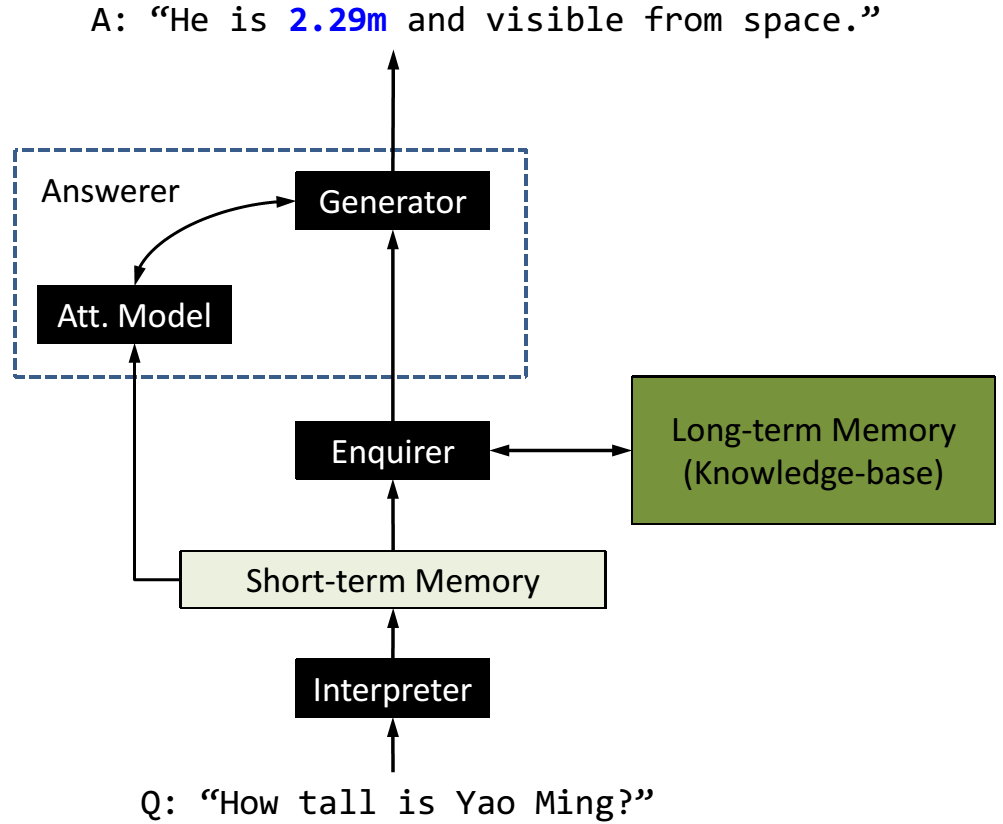}
		\end{tabular}
		\caption{System diagram of \textsc{genQA}.} 
		\label{fig:diagram}
	\end{center}
\end{figure}

Let $Q=(x_1, \ldots, x_{T_Q})$ and $Y=(y_1, \ldots, y_{T_Y})$ denote the natural language question and answer respectively.
The knowledge-base is organized as a set of triples (\textit{subject}, \textit{predicate}, \textit{object}), each denoted as $\tau =(\tau_s, \tau_p, \tau_o)$.
Inspired by the work on the encoder-decoder framework for neural machine translation~\cite{cho2014learning,sutskever2014sequence,bahdanau2014neural} and neural natural language dialogue~\cite{shang15neural,vinyals2015neural,serban2015building}, and the work on question answering with knowledge-base embedding~\cite{bordes2014open,bordes2014question,bordes2015large}, we propose an end-to-end neural network model for generative QA, named \textsc{genQA}, which is illustrated in Figure~\ref{fig:diagram}.

The \textsc{genQA} model consists of {\bf Interpreter}, {\bf Enquirer}, {\bf Answerer}, and an external knowledge-base. {\bf Answerer} further consists of {\bf Attention Model} and {\bf Generator}. Basically,  {\bf Interpreter} transforms the natural language question $Q$ into a representation $\HH_Q$ and saves it in the {short-term memory}. {\bf Enquirer} takes $\HH_Q$ as input to interact with the knowledge-base in the long-term memory, retrieves relevant facts (triples) from the knowledge-base, and summarizes the result in a vector $\rr_{Q}$. The {\bf Answerer} feeds on the question representation $\HH_Q$ (through the {\bf Attention Model}) as well as the vector $\rr_{Q}$ and generates an answer with {\bf Generator}. We elaborate each component hereafter.

\subsection{Interpreter}
Given the question represented as word sequence $Q=(x_1, \ldots, x_{T_Q})$, Interpreter encodes it to an array of vector representations. In our implementation, we adopt a bi-directional recurrent neural network (RNN) as in~\cite{bahdanau2014neural}, which processes the sequence forward and backward by using two independent RNNs (here we use gated recurrent unit (GRU)~\cite{chung2014empirical}). By concatenating the hidden states (denoted as $(\h_{1},\cdots,\h_{T_Q})$), the embeddings of words  (denoted as $(\x_{1},\cdots,\x_{T_Q})$), and the one-hot representations of words, we obtain an array of vectors $\HH_Q = (\tilde{\h}_{1},\cdots,\tilde{\h}_{T_Q})$, where $\tilde{\h}_t = [\h_t; \x_t; x_t]$. This array of vectors is saved in the short-term memory, allowing for further processing by Enquirer and Answerer.

\subsection{Enquirer}
\begin{figure}
	\begin{center}
		\begin{tabular}[c]{cc}
			\includegraphics[width=0.65\textwidth]{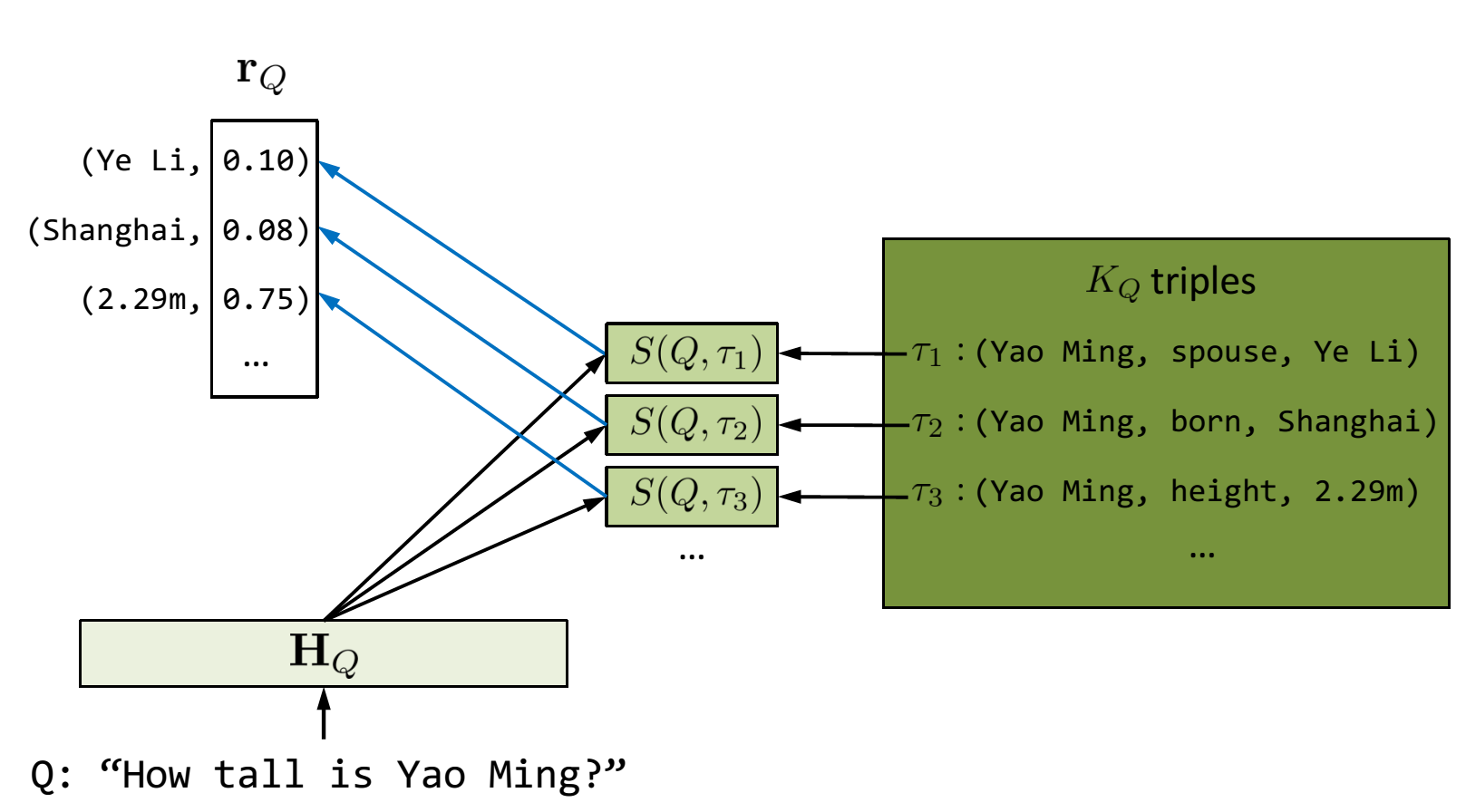}
		\end{tabular}
		\caption{Enquirer of \textsc{genQA}.} 
		\label{f:enqurier}
	\end{center}
\end{figure}

Enquirer ``fetches" relevant facts from the knowledge-base with $\HH_Q$ (as illustrated in Figure~\ref{f:enqurier}). Enquirer first performs term-level matching (similar to the method of associating question-answer pairs with triples described in Section~\ref{sec:learning_task}) to retrieve a list of relevant candidate triples, denoted as $\calT_Q=\{\tau_k\}_{k=1}^{K_Q}$. $K_Q$ is the number of candidate triples, which is at most several hundreds in our data. After obtaining $\calT_Q$,  Enquirer then evaluates the relevance of each candidate triple with the question in the embedded space\cite{bordes2014open,bordes2014question}.

More specifically Enquirer calculates the relevance (matching) scores between the question and the $K_Q$ triples. For question $Q$, the scores are represented in a $K_Q$-dimensional vector $\rr_Q$ where the $k^{th}$ element of $\rr_Q$ is defined as the probability
\[
{r_{Q}}_k = \frac{e^{S(Q, \tau_k)}}{\sum_{k'=1}^{K_Q}e^{S(Q, \tau_{k'})}},
\]
where $S(Q, \tau_k)$ denotes the matching score between question $Q$ and triple $\tau_k$.

The probability in $\rr_Q$ will be further taken into the probabilistic model in Answerer for generating an answer. Since $\rr_Q$ is of modest size, the number of triples involved in the matching score calculation is limited and the efficiency of the process is significantly enhanced. This is particularly true in the learning phase in which the parameters can be efficiently optimized from the supervision signals through back-propagation.

In this work, we provide two implementations for Enquirer to calculate the matching scores between question and triples.

\paragraph{Bilinear Model}
The first implementation simply takes the average of the word embedding vectors in $\HH_Q$ as the representation of the question (with the result denoted as $\bar{\x}_Q$) . For each triple $\tau$ in the knowledge-base, it takes the average of the embeddings of its \textit{subject} and \textit{predicate} as the representation of the triple (denoted as $\uu_\tau$). Then we define the matching score as
\[
\bar{S}(Q, \tau)=\bar{\x}_{Q}^\top \M \uu_\tau,
\]
where $\M$ is a matrix parameterizing the matching between the question and the triple.
\paragraph{CNN-based Matching Model}
The second implementation employs a convolutional neural network (CNN) for modeling the matching score between the question and the triple, as in~\cite{hu2014convolutional} and~\cite{shen2014learning}. Specifically, the question is fed to a convolutional layer followed by a max-pooling layer, and summarized as a fixed-length vector, denoted as $\hat{\h}_{Q}$. Then $\hat{\h}_{Q}$ and $\uu_\tau$ (again as the average of the embedding of the corresponding \textit{subject} and \textit{predicate}) are concatenated as input to a multi-layer perceptron (MLP) to produce their matching score
\[
\hat{S}(Q, \tau)=f_{\textrm{MLP}}([\hat{\h}_{Q};\uu_\tau]).
\]
For this model the parameters consist of those in the CNN and the MLP.

\subsection{Answerer}
\begin{figure}
	\begin{center}
		\begin{tabular}[c]{cc}
			\includegraphics[width=0.5\textwidth]{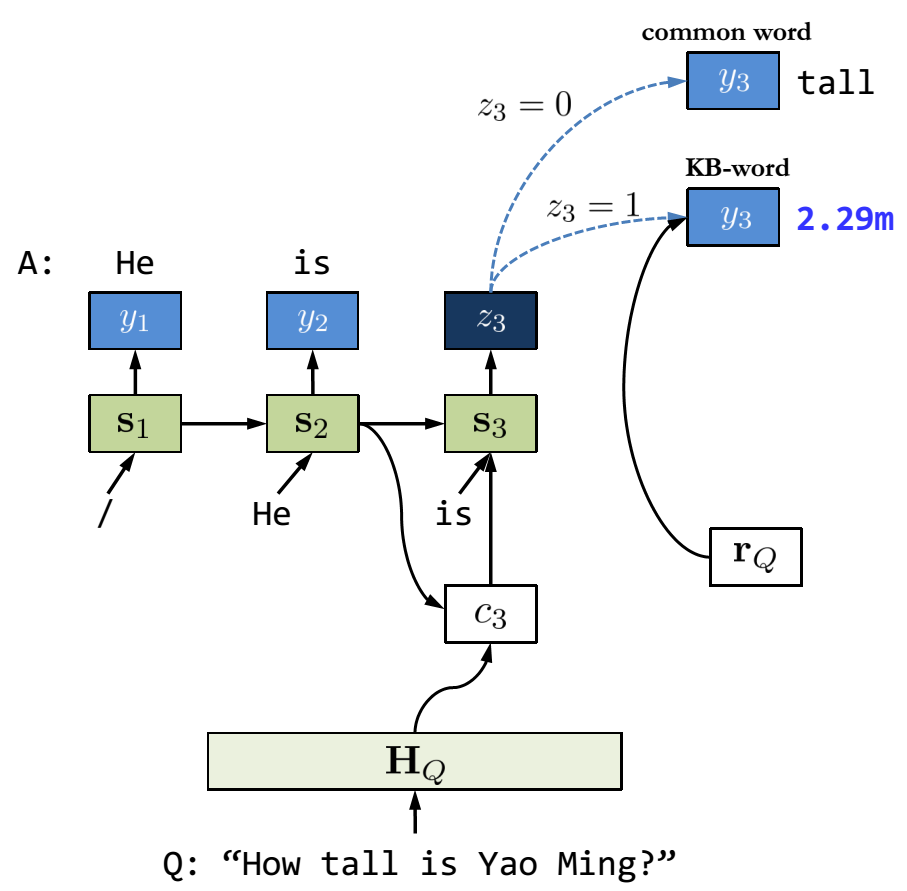}
		\end{tabular}
		\caption{Answerer of \textsc{genQA}.} 
		\label{f:answerer}
	\end{center}
\end{figure}

Answerer uses an RNN to generate an answer based on the information of question saved in the short-term memory (represented as $\HH_Q$) and the relevant facts retrieved from the long-term memory (indexed by $\rr_Q$), as illustrated in Figure~\ref{f:answerer}. The probability of generating the answer $Y=(y_1, y_2,\ldots, y_{T_Y})$ is defined as
\[
\begin{split}
p(y_1,\cdots,& y_{T_Y} | \HH_Q, \rr_Q; \theta)  = p(y_1|\HH_Q, \rr_Q; \theta) \prod_{t=2}^{T_Y}p(y_t|y_1,\ldots,y_{t-1}, \HH_Q, \rr_Q; \theta)
\end{split}
\]
where $\theta$ represents the parameters in the \textsc{genQA} model. The conditional probability in the RNN model (with hidden states $\sss_1,\cdots,\sss_{T_Y}$) is specified by
\[
p(y_t|y_1,\ldots,y_{t-1}, \HH_Q, \rr_Q; \theta) = p(y_t|y_{t-1}, \sss_t, \HH_Q, \rr_Q; \theta).
\]
In generating the $t^{th}$ word $y_t$ in the answer, the probability is given by the following mixture model
\[
\begin{split}
p(y_t|y_{t-1}, & \sss_t, \HH_Q, \rr_Q; \theta) = \\
& p(z_t=0|\sss_t; \theta)p(y_t|y_{t-1}, \sss_t, \HH_Q, z_t=0; \theta) + p(z_t=1|\sss_t; \theta)p(y_t|\rr_Q, z_t=1; \theta),
\end{split}
\]
which sums the contributions from the ``language" part and the ``knowledge'' part, with the coefficient $p(z_t|\sss_t; \theta)$ being realized by a logistic regression model with $\sss_t$ as input. Here the latent variable $z_t$ indicates whether the $t^{th}$ word is generated from a common vocabulary (for $z_t = 0$) or a KB vocabulary ($z_t = 1$). In this work, the KB vocabulary contains all the \textit{objects} of the candidate triples associated with the particular question. For any word $y$ that is \emph{only} in the KB vocabulary, e.g., ``\texttt{2.29m}", we have $p(y_t|y_{t-1}, \sss_t, \HH_Q, z_t=0; \theta) = 0$, while for $y$ that does not appear in KB, e.g., ``\texttt{and}", we have $p(y_t|\rr_Q, z_t=1; \theta) = 0$. There are some words (e.g., ``\texttt{Shanghai}") that appear in both common vocabulary and KB vocabulary, for which the probability contains nontrivial contributions from both bodies.

In generating common words, Answerer acts in the same way as the decoder of RNN in~\cite{bahdanau2014neural} with information from $\HH_Q$ selected by the {attention model}. Specifically, the hidden state at $t$ step is computed as $\sss_t=f_s(y_{t-1}, \sss_{t-1}, c_{t})$ and $p(y_t|y_{t-1}, \sss_t, \HH_Q, z_t=0; \theta) = f_y(y_{t-1}, \sss_t, c_{t})$, where $c_{t}$ is the context vector computed as a weighted sum of the hidden states stored in the short-term memory $\HH_Q$.

In generating KB-words via $p(y_t|\rr_Q, z_t=1; \theta)$, Answerer simply employs the model $p(y_t=k|\rr_Q, z_t=1; \theta)={r_{Q}}_k$. The better a triple matched with the question, the more likely the \textit{object} of the triple is selected.

\subsection{Training}
The parameters to be learned include the weights in the RNNs for Interpreter and Answerer, parameters in Enquirer (either the matrix $\M$ or the weights in the convolution layer and MLP), and the word-embeddings which are shared by the Interpreter RNN and the knowledge-base. \textsc{genQA}, although essentially containing a retrieval operation, can be trained in an end-to-end fashion by maximizing the likelihood of observed data, since the mixture form of probability in Answerer provides a unified way to generate words from the common vocabulary and the KB vocabulary. Specifically, given the training data $\calD = \{(Q^{(i)}, Y^{(i)}, \calT_Q^{(i)})\}$, the optimal parameters are obtained by minimizing the negative log-likelihood with regularization on all the parameters
\[
\ell(\calD, \theta) = -\sum_{i=1}^{N_{\calD}} \log p(Y^{(i)}|Q^{(i)}, \calT_Q^{(i)}) + \lambda \lVert \theta \rVert_F^2.
\]
In practice the model is trained on machines with GPUs by using stochastic gradient-descent with mini-batch.

\section{Experiments}\label{sec:experiments}

\subsection{Implementation details}
The texts in Chinese in the data are converted into sequences of words using the Jieba Chinese word segmentor. Since the word distributions on questions and answers are different, we use different vocabularies for them. Specifically for questions, we use the most frequent 30K words in the questions and all the words in the \textit{predicates} of the triples, covering 98.4\% of the word usages in the questions. For answers, we use the most frequent 30K words in the answers with a coverage of 97.3\%. All the out-of-vocabulary words are replaced by a special token ``UNK''. The dimensions of the hidden states of encoder and decoder are both set to 500, and the dimension of the word-embedding is set to 300. Our models are trained on an NVIDIA Tesla K40 GPU using Theano~\cite{bastien2012theano,bergstra2010theano}, with the mini-batch size of 80. The training of each model takes about two or three days.

\subsection{Comparison Models}

To our best knowledge there is no previous work on generative QA, we choose three baseline methods: a neural dialogue model, a retrieval-based QA model, and an embedding based QA model, respectively corresponding to the generative aspect and the KB-retrieval aspect of \textsc{genQA}:\\

\noindent \textbf{Neural Responding Machine (NRM)}:  NRM~\cite{shang15neural} is a neural network based generative model specially designed for short-text conversation. We train the NRM model with the question-answer pairs in the training data having the same vocabulary as  \textsc{genQA}. Since NRM does not access the knowledge-base during training and test, it actually remembers all the knowledge from the QA pairs in the model.\\

\noindent \textbf{Retrieval-based QA}: the knowledge-base is indexed by an information retrieval system (we use Apache Solr), in which each triple is deemed as a document. At the test phase, a question is used as the query and the top-retrieved triple is returned as the answer. Note that this method cannot generate natural language answers.\\

\noindent \textbf{Embedding-based QA}: as proposed by~\cite{bordes2014question,bordes2014open}, the model is learnt from the question-triple pairs in the training data. The model learns to map questions and knowledge-base constituents into the same embedding space, where the similarity between question and triple is computed as the inner product of two embedding vectors. Different from the cross-entropy loss used in \textsc{genQA}, this model uses a ranking loss function as follows:
\[
\sum_{i=1}^{N_{\calD}}\sum_{\tau,\tau'\in \calT^{(i)}} \max(0, m-S(Q^{(i)},\tau)+S(Q^{(i)},\tau')),
\]
where $\tau$ and $\tau'$ represent the positive and negative triples corresponding to the question.  Similar to the retrieval-based QA, this model cannot generate natural language answers either.

Since we have two implementations of Enquirer of the ~\textsc{genQA} model, we denote the one using the bilinear model as \textsc{genQA} and the other using CNN and MLP as $\textsc{genQA}_\textsf{CNN}$.

\subsection{Results}
We evaluate the performance of the models in terms of 1) accuracy, i.e., the ratio of correctly answered questions, and 2) the fluency of answers. In order to ensure an accurate evaluation, we randomly select 300 questions from the test set, and manually remove the nearly duplicate cases and filter out the mistaken cases (e.g., non-factoid questions).

\paragraph{Accuracy}
Table~\ref{tab:result} shows the accuracies of the models on the test set. NRM has the lowest accuracy, showing the lack of ability to accurately remember the answers and generalize to questions unseen in the training data. For example, to question ``\texttt{\small Which country does Xavi play for as a midfielder?}" (Translated from Chinese), NRM gives the wrong answer ``\texttt{\small He plays for France}" (Translated from Chinese), since the athlete actually plays for Spain. The retrieval-based method achieves a moderate accuracy, but like most string-matching methods it suffers from word mismatch between the question and the triples in the KB. The embedding-based QA model achieves higher accuracy on test set, thanks to  its generalization ability from distributed representations. \textsc{genQA} and $\textsc{genQA}_\textsf{CNN}$ are both better than the competitors, showing that \textsc{genQA} can further benefit from the end-to-end training of sequence-to-sequence learning.  We conjecture that the task of generating the appropriate answers may help the learning of word-embeddings of questions. Among the two \textsc{genQA} variants, $\textsc{genQA}_\textsf{CNN}$  achieves the best accuracy, getting over half of the questions right. An explanation for that is that the convolution layer helps to capture salient features in matching. The experiment results demonstrate the ability of \textsc{genQA} models to find the right answers from the KB even with regard to new facts. For example, to the example question mentioned above, \textsc{genQA} gives the correct answer ``\texttt{\small He plays for Spain}".

\begin{table}
	\centering
	\caption{Test accuracies}\label{tab:result}
	\begin{tabular}{|l|c|}
		\hline
		\textbf{Models} & \textbf{Test} \\
		\hline
		Retrieval-based QA & 36\%   \\
		\hline
		NRM\textsuperscript{\cite{shang15neural}} & 19\%   \\
		\hline
		Embedding-based QA \textsuperscript{\cite{bordes2014open}}  & 45\%   \\
		\hline
		\textsc{genQA} & 47\%  \\
		\hline
		$\textsc{genQA}_\textsf{CNN}$ & \textbf{52}\%  \\
		\hline
	\end{tabular}
\end{table}

\begin{figure*}
	\begin{center}
		\caption{Examples of the generated answers by \textsc{genQA}.}\label{f:examples}
		\includegraphics[width=0.85\textwidth]{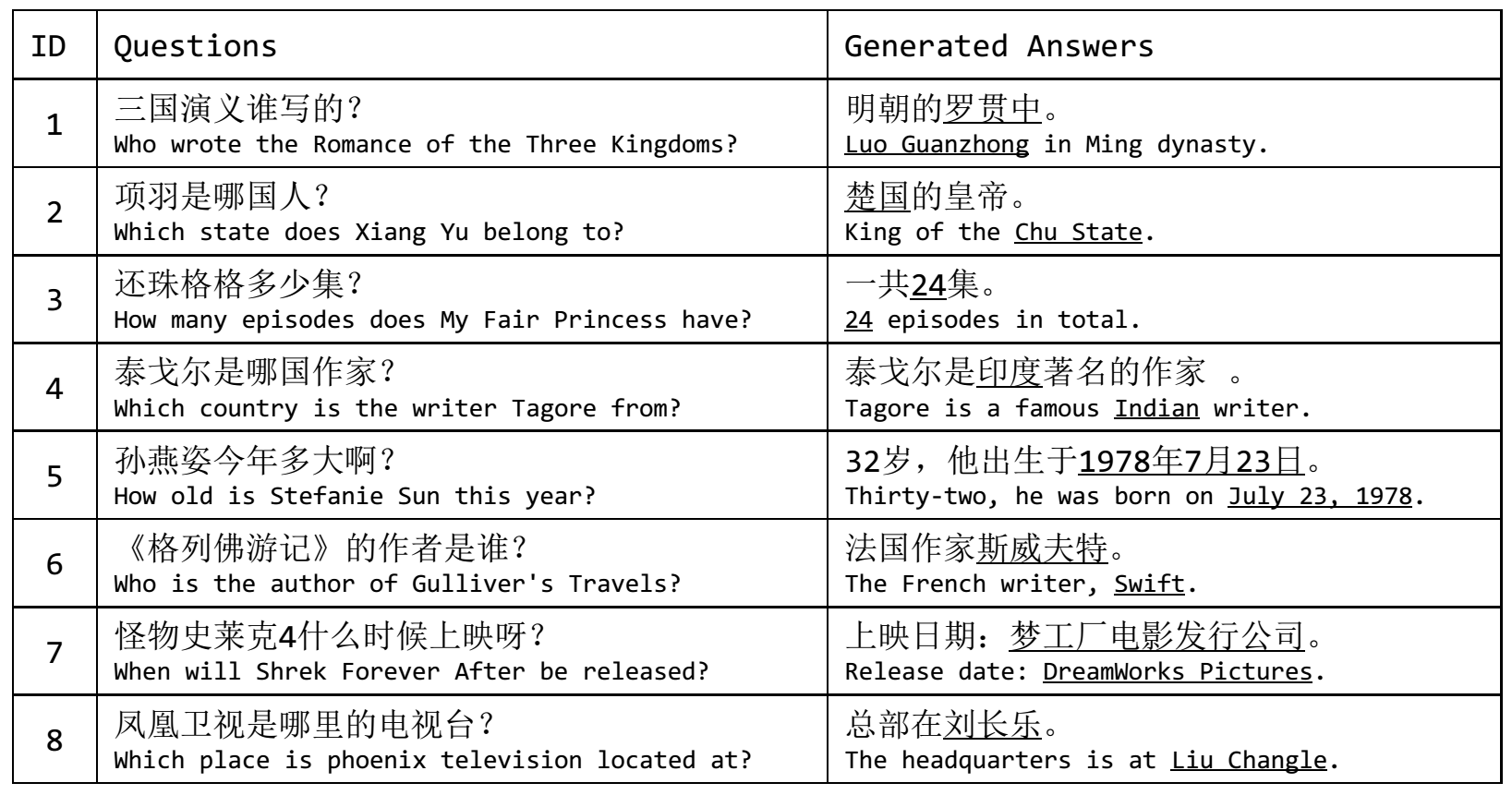}
	\end{center}
\end{figure*}

\paragraph{Fluency}
We make some empirical comparisons and find no significant differences between NRM and \textsc{genQA} in terms of the fluency of answers. In general, all the three generation-based models (two of them are ours) yield correct sentences in most of the time.

\subsection{Case Study}
Figure~\ref{f:examples} gives some examples of generated answers to the questions in the test set by our \textsc{genQA} models, with the underlined words generated from KB. Clearly it can smoothly blend KB-words and common words in the sentences, thanks to the unified neural model that can learn to determine the right time to place a KB-word or a common word. We notice that most of the generated answers are short sentences, for which there are two possible reasons:  1) many answers to the factoid questions in the dataset are usually short, and 2) we select the answer for each question by beam-searching the sequence with maximum log-likelihood normalized by its length, which generally prefers a short answer. Examples 1 to 4 show the correctly generated answers, for which the model not only matches the right triples (and thus generates the right KB-words), but also generates suitable common words surrounding them. However, in some cases like examples 5 and 6 even the right triples are found, the surrounding common words are improper or incorrect from the knowledge-base point of view (e.g., in example 6 the author ``\texttt{Jonathan Swift}" is from Ireland rather than France). By investigating the correctly generated answers on test data, we find that roughly 8\% of them having improper surrounding words. In some other cases, the model fails to match the correct triples with the questions, which produces completely wrong answers. For example 7, the question is about the release date of a movie, while the model finds its distributor and generates an answer incorrect both in terms of fact and language.

\section{Related Work}
Our work is inspired by recent work on neural machine translation and neural natural language dialogue. Most of neural translation models fall into the encoder-decoder framework~\cite{cho2014learning,cho2014properties,sutskever2014sequence}, where the encoder summarizes the input sequence into a sequence of vector representations and the decoder generates the output sequence from the sequence of vector representations. Bahdanau et al.~\cite{bahdanau2014neural} introduce the attention mechanism into the framework, and their system known as RNNsearch algorithm can jointly learn alignment and translation, and significantly improve the translation quality. This framework has also been used in natural language dialogue~\cite{shang15neural,vinyals2015neural,serban2015building,wen2015semantically,wen2015stochastic}, where the end-to-end neural dialogue model is trained on a large amount of conversation data.  Although promising, neural dialogue models still have problems and limitations, e.g., the lack of mechanism to incorporate knowledge.

Our work is also inspired by recent work on knowledge-base embedding and question answering from knowledge-base. TransE~\cite{bordes2013translating} is a method that learns the embedding vectors of the entities and the relations between entities by translating from \textit{subject} entities to \textit{object} entities. The model for question answering learns to embed questions and constituents in knowledge-base in the same low-dimensional space, where the similarity score between a question and a triple/subgraph is computed and the top ranked triples/subgraphs are selected as answers~\cite{bordes2014open,bordes2014question}. Yang et al.~\cite{yang2014joint} propose a method that transforms natural questions into their corresponding logical forms using joint relational embeddings, and conducts question answering
by leveraging semantic associations between lexical representations and KB properties in the latent space.

Memory Networks~\cite{memorynet,sukhbaatar2015end} is a recently proposed class of models that combine a large memory with a learning component that can read and write to the memory, in order to conduct reasoning for QA. Bordes et al.~\cite{bordes2015large} present an embedding-based question answering system developed under the framework of memory networks, which shows the perspective to involve more inference schemes in QA. Recently, Yin et al.~\cite{yin2015neural} propose an architecture, known as Neural Enquirer, to execute a natural language query on knowledge-base tables for question answering. It is a fully neural and end-to-end network that uses distributional representations of the query and the table, and realizes the execution of a compositional query through a series of differentiable operations.

\section{Conclusion}\label{sec:conclusion}
In this paper we have proposed an end-to-end neural network model for generative question answering. The model is built on the encoder-decoder framework for sequence-to-sequence learning, while equipped with the ability to query a knowledge-base. Empirical studies show the proposed model is capable of generating natural and right answers to the questions by referring to the facts in the knowledgebase. In the future, we plan to continue the work on question answering and dialogue, which includes: 1) iterative question answering: a QA system that can interact with the user to confirm/clarify/answer her questions in a multi-turn dialogue; 2) question answering from complex knowledgebase: a QA system that has the ability of querying a complex-structured knowledge-base such as a knowledge graph.

\bibliographystyle{abbrv}
\small \bibliography{GenQA}

\end{document}